# Do We Need a Detailed Rubric for Automated Essay Scoring using Large Language Models?

Lui Yoshida

The University of Tokyo, Tokyo, Japan
`luiyoshida@g.ecc.u-tokyo.ac.jp`

**Abstract.** This study investigates the necessity and impact of a detailed rubric in automated essay scoring (AES) using large language models (LLMs). While using rubrics are standard in LLM-based AES, creating detailed rubrics requires substantial effort and increases token usage. We examined how different levels of rubric detail affect scoring accuracy across multiple LLMs using the TOEFL11 dataset. Our experiments compared three conditions: a full rubric, a simplified rubric, and no rubric, using four different LLMs (Claude 3.5 Haiku, Gemini 1.5 Flash, GPT-4o-mini, and Llama 3 70B Instruct). Results showed that three out of four models maintained similar scoring accuracy with the simplified rubric compared to the detailed one, while significantly reducing token usage. However, one model (Gemini 1.5 Flash) showed decreased performance with more detailed rubrics. The findings suggest that simplified rubrics may be sufficient for most LLM-based AES applications, offering a more efficient alternative without compromising scoring accuracy. However, model-specific evaluation remains crucial as performance patterns vary across different LLMs.

## 1    Introduction

The development of Large Language Models (LLMs) has marked a significant breakthrough in artificial intelligence, showing remarkable progress and versatility across various fields [1-5]. These advances have made substantial impacts in education, where LLMs are being actively adopted and tested in different learning contexts [6-8]. A notable application in this domain is automated essay scoring (AES), which enables automatic evaluation of essays. AES represents a well-established research field with over fifty years of continuous development and improvement [9-12]. In recent years, fine-tuned deep neural networks, especially those based on BERT architectures, have shown superior performance in this task, setting new standards for automated assessment accuracy.

The application of LLMs in AES has gained significant attention from researchers worldwide [13-19]. For instance, Lee et al. (2024) [18] achieved significant performance improvements over existing methods by automatically generating multiple evaluation criteria and having LLMs perform specialized assessments for each criterion, thereby enhancing both evaluation accuracy and consistency. In such research using LLMs for AES, including rubrics as evaluation criteria in the prompt has become a standard approach. It is well known that proper utilization of rubrics in evaluation can enhance validity and reliability [20-22]. However, creating detailed rubrics requires substantial effort and expertise [23-25], which poses a challenge for implementation in educational settings. Additionally, in the context of LLM-based AES, including detailed rubrics in prompts increases the required token count. When using LLMs, not only do costs increase with token usage, but the environmental impact of model execution must also be considered [26,27]. Therefore, finding more token-efficient methods while maintaining scoring accuracy is an important challenge.

This study aims to examine the necessity of rubrics in LLM-based AES and investigate how their level of detail affects scoring accuracy. Specifically, we seek to answer research questions such as "Are detailed rubrics really necessary?" "Can simplified rubrics maintain sufficient scoring accuracy?" and "Are rubrics necessary at all?" Furthermore, given the recent emergence of multiple high-performance LLMs, we will also clarify how robust these trends are across different models.

## 2      Methods

**Dataset.** We used TOEFL11 [28] as the essay dataset. The dataset contains eight essay prompts and their corresponding examinee essays, each with approximately 1,000 to 1,600 essays, for a total of 12,100 essays. The data also includes expert ratings of the essays on a three-point scale of high, medium, and low. These ratings were first evaluated by several experts using a 5-point rubric and finally compressed to a 3-point rating according to a set of rules. The rubric ratings are not included in the dataset. We used all essays for AES.

**Large Language Models.** To evaluate the effects across multiple model types, we conducted AES using Anthropic's Claude 3.5 Haiku (claude-3-5-haiku-20241022), OpenAI's GPT-4o-mini (gpt-4o-mini-2024-07-18), Google's Gemini 1.5 Flash (gemini-1.5-flash), and Meta's Llama 3 70B Instruct (llama-3-70b-instruct). We accessed these models through their respective APIs: Anthropic's API for Claude, Microsoft Azure OpenAI Service API for GPT, Google's API for Gemini, and Replicate's API for Llama.

**Prompts.** We developed prompts based on those used by Yancey et al. [14]. The prompts comprised several components: Instruction, Essay Prompt, Response,  Assessment Method, and Output Format (**Fig. 1**). The Instruction section described the essay evaluation task, while the Essay Prompt section presented the prompt for the essay. The Response section contained the essay to be evaluated. The Assessment Method varied depending on conditions as described later. Lastly, the Output Format section was the template for output.

In this study, we prepared three different Assessment Method prompts. One was a prompt that included a detailed rubric for evaluation (denoted as Full Rubric), another was a prompt that included a simplified rubric (denoted as Simplified Rubric), and the You are a rater for writing responses on a high-stakes English language exam for second language learners. You will be provided with a prompt and the test-taker's response. Your rating should be based on the rubric below, following the specified format. There are rating samples of experts so that you can refer to those when rating.

\# Prompt
"""*Essay prompt*"""

\# Response
"""*Essay to be evaluated*"""

\# Assessment Method
*Assessment Method*

\# Output format:
Rating: [<<<Your rating here.>>>]

**Fig. 1.** A template of a prompt. The parts where data should be inserted are in *italics*.

third was a prompt without any rubric (denoted as None). For the detailed rubric, we used the actual rubric used in the TOEFL 11 evaluation [28]. For the simplified rubric, we used a version (**Fig. 2**) that the authors simplified based on the original rubric. In the prompt without a rubric, we simply stated "Rate the response on a scale of 0 to 5."

**Agreement between Experts and LLMs.** To evaluate the agreement between experts and LLMs, we used the QWK, a widely used metric to assess the concordance between machine and human evaluations [11,29]. Since evaluations of LLMs used a rubric out of 5 points or a 5-point scale, we converted these into three levels: scores above 3 as high, 3 as medium, and below 3 as low, and numerically coded these as 3, 2, 1, respectively, to calculate the QWK according to the original method [28].

We calculated QWK for each combination of model and prompt. Additionally, we performed 1,000 bootstrap resampling iterations to determine the 95% confidence intervals. Furthermore, to examine the differences in QWK between prompts within each model, we conducted significance tests at the 5% level using paired bootstrap with 1,000 resampling iterations. Given the multiple comparisons being made, we applied Holm's correction to adjust the p-values.

To understand the relationship between expert evaluations and LLM evaluations, we also created confusion matrices showing the distribution of scores assigned by LLMs relative to the experts' scores.

**Token count.** To evaluate how input token counts differed across prompt types, we calculated the mean and standard deviation of input prompt tokens across all essays for each model. For the Llama models, we excluded them from the results as their API specifications did not allow us to obtain input token counts.

5: Fully addresses the topic with clear organization, strong examples, smooth flow, and excellent language use.
4: Addresses the topic well, with good organization and examples, though some points could be clearer.
3: Covers the topic but lacks depth or clarity, with limited language variety and occasional errors.
2: Weak response with poor organization, insufficient examples, and frequent language errors.
1: Very weak, disorganized, or off-topic with serious language issues.
0: Off-topic, copied, in another language, or blank.

**Fig. 2.** A simplified rubric (78 words). It was based on an original rubric (375 words) [16].

## 3      Results

The QWK calculation results are shown in **Fig. 3**. While Claude and Llama showed significantly higher QWK with Full Rubric compared to Simplified Rubric, the difference was less than 0.01, indicating minimal practical difference. For GPT, no significant difference in QWK was observed between Full Rubric and Simplified Rubric. In these three models, prompts containing rubrics showed significantly higher QWK than those without rubrics. In contrast, Gemini exhibited a trend where QWK significantly decreased as rubric detail increased.

Confusion matrices showing the relationship between expert evaluations and LLM evaluations are presented in **Fig. 4**. For Gemini, which showed decreasing QWK with more detailed rubrics, there was a tendency to assign higher scores as rubric became more detailed, leading to divergence from expert evaluations. However, other models did not show similar trends, indicating that the impact of rubric detail varied depending on the model.

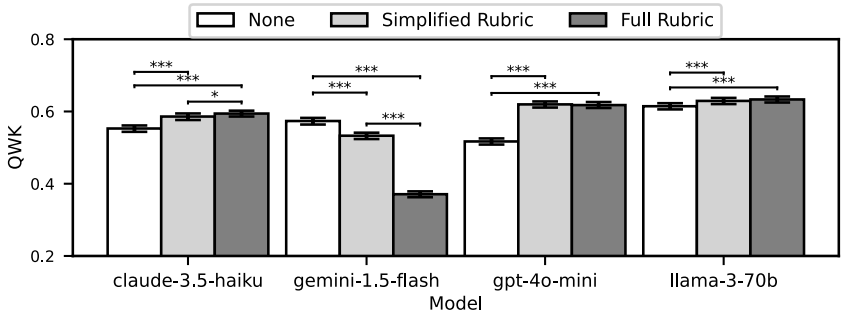

**Fig. 3.** QWK between expert ratings and AI ratings (\*: p<0.05, \*\*\*: p<0.001). Error bars indicate 95% confidence intervals.

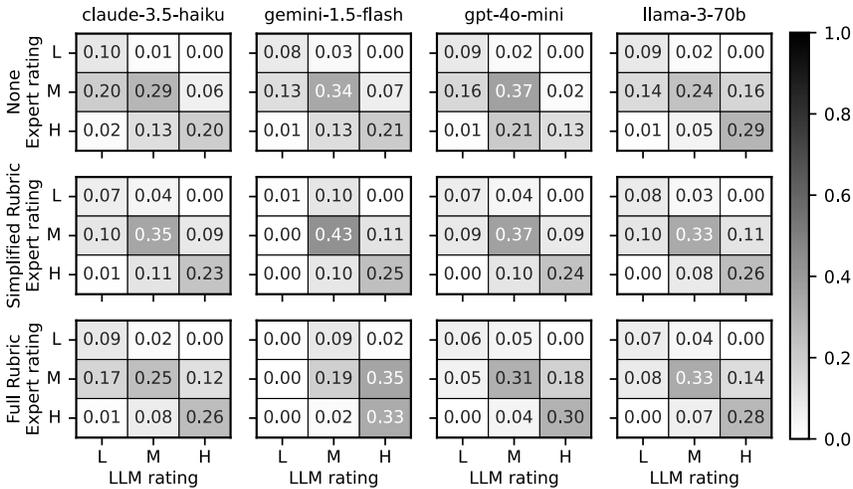

**Fig. 4.** Confusion matrices showing the correspondence between expert and LLM ratings. Cell values show percentages of total cases. L, M, H represent low, medium, high.

**Table 1.** Average token count of all prompts for each type with standard deviation (std).

| Model | None | Simplified Rubric | Full Rubric | std. |
|-------|------|-------------------|-------------|------|
| Claude 3.5 Haiku | 533.4 | 665.4 | 1077.4 | 91.1 |
| Gemini 1.5 Flash | 516.4 | 640.4 | 1011.4 | 88.1 |
| GPT 4o-mini | 501.8 | 618.8 | 992.8 | 88.3 |

The mean input token counts and standard deviations for each prompt across all essays are shown in **Table 1**. While Full Rubric prompts required approximately twice as many tokens compared to prompts without rubrics, Simplified Rubric prompts required only about 1.2 times as many tokens, demonstrating better token efficiency.

## 4    Discussion

Among three out of the four models, there was no significant difference in evaluation performance between Simplified Rubric and the Full Rubric. However, with Gemini 1.5 Flash, it was observed that performance decreased as the rubric became more detailed. It is known that for some models, longer prompts can lead to decreased inference performance or reduced adherence to the prompt [30,31]. This phenomenon may have occurred with Gemini 1.5 Flash. In this context, even within the same model series, different results might be obtained with higher-performing models.

Next, we discuss implications for practical applications. Since performance does not significantly decrease with Simplified Rubric for three models, detailed rubrics may not be necessary when using them as evaluation aids. Additionally, simplified rubrics are more efficient in terms of both development effort and token count. However, with QWK around 0.6, and considering that QWK between experts is typically around 0.8-0.9 [14], it can be said that AES using LLMs with prompting alone, as implemented in the widely used models examined here, is not yet ready for practical use. For practical applications, while sacrificing some versatility, improvement in essay evaluation capabilities specific to the data can be expected through techniques like fine-tuning [32-34], which have shown performance improvements. However, caution is needed as some models, like Gemini 1.5 Flash, may show decreased performance when given rubrics, making it essential to evaluate each model's assessment capabilities individually. To summarize the practical implications: (1) Since performance does not decrease with the simplified rubric in three out of four models, using a simplified rubric is advantageous both in terms of effort and cost. (2) One model shows decreased performance with more detailed rubrics, necessitating separate evaluation for each model. (3) With QWK around 0.6, these models do not possess expert-level evaluation capabilities, making it crucial not to overly rely on LLM outputs.

Finally, we discuss the limitations of this study. Since the effects on models other than those used in this research remain unclear, there is a need to verify our findings with other models. Additionally, as recently released reasoning models [5,35,36] may exhibit different behaviors, there is room for further investigation. Furthermore, while this study only used the TOEFL11 dataset, we expect that verifying our findings with other datasets such as the Automated Student Assessment Prize (ASAP) program data [37] or the Cambridge Learner Corpus-First Certificate in English Exam (CLC-FCE) [38] would enable greater generalization of our insights.

## 5      Conclusion

This study examined the impact of rubric detail on LLM-based AES performance, yielding several key findings. First, for three of the four tested models, the simplified rubric achieved comparable performance to the detailed rubric while significantly reducing token usage. This suggests that detailed rubrics may not be necessary for effective LLM-based essay scoring in most cases. Second, model-specific variations were observed, with Gemini 1.5 Flash showing decreased performance with more detailed rubrics, highlighting the importance of individual model evaluation. Third, while LLM-based scoring showed promise, the achieved QWK scores around 0.6 indicate that current implementations may not yet match human expert-level performance (typically 0.8-0.9).

**Acknowledgment.** In preparing this manuscript, we used Claude Pro and ChatGPT Pro to refine the language and generate examples of Python code. The tools did not contribute to generating any original ideas. This work was supported by JSPS KAKENHI Grant Number 23K02707.

## References

1. Brown, T., Mann, B., Ryder, N., Subbiah, M., Kaplan, J.D., Dhariwal, P., Neelakantan, A., Shyam, P., Sastry, G., Askell, A., Agarwal, S., Herbert-Voss, A., Krueger, G., Henighan, T., Child, R., Ramesh, A., Ziegler, D., Wu, J., Winter, C., Hesse, C., Chen, M., Sigler, E., Litwin, M., Gray, S., Chess, B., Clark, J., Berner, C., McCandlish, S., Radford, A., Sutskever, I., Amodei, D.: Language Models are Few-Shot Learners. In: Larochelle, H., Ranzato, M., Hadsell, R., Balcan, M.F., and Lin, H. (eds.) Advances in Neural Information Processing Systems. pp. 1877–1901. Curran Associates, Inc. (2020)
2. Wei, J., Wang, X., Schuurmans, D., Bosma, M., Ichter, B., Xia, F., Chi, E., Le, Q., Zhou, D.: Chain-of-Thought Prompting Elicits Reasoning in Large Language Models. In: Koyejo, S., Mohamed, S., Agarwal, D., Belgrave, D., Cho, K., and Oh, A. (eds.) Advances in Neural Information Processing Systems. pp. 24824-24837. Curran Associates, Inc. (2022)
3. Kojima, T., Gu, S.S., Reid, M., Matsuo, Y., Iwasawa, Y.: Large Language Models are Zero-Shot Reasoners. In: Koyejo, S., Mohamed, S., Agarwal, D., Belgrave, D., Cho, K., and Oh, A. (eds.) Advances in Neural Information Processing Systems. pp. 22199—22213. Curran Associates, Inc. (2022)
4. OpenAI: GPT-4 Technical Report, http://arxiv.org/abs/2303.08774, (2023)
5. OpenAI: OpenAI o1 System Card, https://openai.com/index/openai-o1-system-card/, last accessed 2025/02/19
6. Kasneci, E., Sessler, K., Küchemann, S., Bannert, M., Dementieva, D., Fischer, F., Gasser, U., Groh, G., Günnemann, S., Hüllermeier, E., Krusche, S., Kutyniok, G., Michaeli, T., Nerdel, C., Pfeffer, J., Poquet, O., Sailer, M., Schmidt, A., Seidel, T., Stadler, M., Weller,

J., Kuhn, J., Kasneci, G.: ChatGPT for good? On opportunities and challenges of large language models for education. Learning and Individual Differences. **103**, 102274 (2023)

7. Yan, L., Sha, L., Zhao, L., Li, Y., Martinez-Maldonado, R., Chen, G., Li, X., Jin, Y., Gašević, D.: Practical and ethical challenges of large language models in education: A systematic scoping review. British Journal of Educational Technology. **55**(1), 90–112 (2024)

8. Jeon, J., Lee, S.: Large language models in education: A focus on the complementary relationship between human teachers and ChatGPT. Education and Information Technologies. **28**, 15873–15892 (2023)

9. Page, E.B.: The Imminence of... Grading Essays by Computer. The Phi Delta Kappan. **47**, 238–243 (1966)

10. Hussein, M.A., Hassan, H., Nassef, M.: Automated language essay scoring systems: a literature review. PeerJ Computer Science. **5**, e208 (2019)

11. Ke, Z., Ng, V.: Automated Essay Scoring: A Survey of the State of the Art. In: Kraus, S. (ed.) Proceedings of the Twenty-Eighth International Joint Conference on Artificial Intelligence. pp. 6300–6308. International Joint Conferences on Artificial Intelligence (2019)

12. Ramesh, D., Sanampudi, S.K.: An automated essay scoring systems: a systematic literature review. Artificial Intelligence Review. **55**, 2495–2527 (2022)

13. Mizumoto, A., Eguchi, M.: Exploring the Potential of Using an AI Language Model for Automated Essay Scoring. Research Methods in Applied Linguistics. **2**(2), 100050 (2023)

14. Yancey, K.P., Laflair, G., Verardi, A., Burstein, J.: Rating Short L2 Essays on the CEFR Scale with GPT-4. In: Kochmar, E., Burstein, J., Horbach, A., Laarmann-Quante, R. Madnani, N., Tack, A., Yaneva, V., Yuan, Z., and Zesch, T. (eds.) Proceedings of the 18th Workshop on Innovative Use of NLP for Building Educational Applications (BEA 2023). pp. 576–584. Association for Computational Linguistics (2023)

15. Naismith, B., Mulcaire, P., Burstein, J.: Automated evaluation of written discourse coherence using GPT-4. In: Kochmar, E., Burstein, J., Horbach, A., Laarmann-Quante, R. Madnani, N., Tack, A., Yaneva, V., Yuan, Z., and Zesch, T. (eds.) Proceedings of the 18th Workshop on Innovative Use of NLP for Building Educational Applications (BEA 2023). pp. 576–584. Association for Computational Linguistics (2023)

16. Kim, S., Jo, M.: Is GPT-4 Alone Sufficient for Automated Essay Scoring?: A Comparative Judgment Approach Based on Rater Cognition. In: Joyner, D., Kim, K.M., Wang, X., and Xia, M. (eds.) Proceedings of the Eleventh ACM Conference on Learning @ Scale. pp. 315–319. Association for Computing Machinery (2024)

17. Yoshida, L.: The Impact of Example Selection in Few-Shot Prompting on Automated Essay Scoring Using GPT Models. In: Olney, A.M., Chounta, I.-A., Liu, Z., Santos, O.C., and Bittencourt, I.I. (eds.) Artificial Intelligence in Education. Posters and Late Breaking Results, Workshops and Tutorials, Industry and Innovation Tracks, Practitioners, Doctoral Consortium and Blue Sky. pp. 61–73. Springer Cham (2024)

18. Lee, S., Cai, Y., Meng, D., Wang, Z., Wu, Y.: Unleashing Large Language Models' Proficiency in Zero-shot Essay Scoring, In: (eds.) Findings of the Association for Computational Linguistics: EMNLP 2024. pp. 181–198. Association for Computational Linguistics (2024)

19. Tate, T.P., Steiss, J., Bailey, D., Graham, S., Moon, Y., Ritchie, D., Tseng, W., Warschauer, M.: Can AI provide useful holistic essay scoring? Computers and Education: Artificial Intelligence. **7**, 100255 (2024)

20. Jonsson A., Svingby G.: The use of scoring rubrics: Reliability, validity and educational consequences. Educational Research Review. **2**(2), 130–144 (2007)

21. Stellmack, M.A., Konheim-Kalkstein, Y.L., Manor, J.E., Massey, A.R., Schmitz, J.A.P.: An Assessment of Reliability and Validity of a Rubric for Grading APA-Style Introductions. Teaching of Psychology. **36**(2), 102–107 (2009)

22. Reddy, Y.M., Andrade, H.: A review of rubric use in higher education. Assessment & Evaluation in Higher Education. **35**(4), 435–448 (2010)
23. Moskal, B.M., Leydens, J.A.: Scoring Rubric Development: Validity and Reliability. Practical Assessment, Research & Evaluation. **7**(1), 10 (2000)
24. Moskal, B.M.: Developing Classroom Performance Assessments and Scoring Rubrics - Part I. ERIC Digest. ERIC Clearinghouse on Assessment and Evaluation College Park MD. ED481714 (2003)
25. Allen, D., Tanner, K.: Rubrics: Tools for Making Learning Goals and Evaluation Criteria Explicit for Both Teachers and Learners. Life Sciences Education. **5**(3), 197–203 (2006)
26. Rillig, M.C., Ågerstrand, M., Bi, M., Gould, K.A., Sauerland, U.: Risks and Benefits of Large Language Models for the Environment. Environmental Science & Technology. **57**(9), 3464–3466 (2023)
27. Argerich, F.M. and Patiño-Martínez, M.: Measuring and Improving the Energy Efficiency of Large Language Models Inference. IEEE Access. **12**, 80194-80207 (2024)
28. Blanchard, D., Tetreault, J., Higgins, D., Cahill, A., Chodorow, M.: TOEFL11: A Corpus of Non-Native English. ETS Research Report. RR-13-24, i–15 (2013)
29. Ramnarain-Seetohul, V., Bassoo, V., Rosunally, Y.: Similarity measures in automated essay scoring systems: A ten-year review. Education and Information Technologies. **27**, 5573–5604 (2022)
30. Levy, M., Jacoby, A., Goldberg, Y.: Same Task, More Tokens: the Impact of Input Length on the Reasoning Performance of Large Language Models. In: Ku, L.-W., Martins, A., and Srikumar, V. (eds.) Proceedings of the 62nd Annual Meeting of the Association for Computational Linguistics (Volume 1: Long Papers). pp. 15339–15353. Association for Computational Linguistics (2024)
31. Liu, N.F., Lin, K., Hewitt, J., Paranjape, A., Bevilacqua, M., Petroni, F., Liang, P.: Lost in the Middle: How Language Models Use Long Contexts. Transactions of the Association for Computational Linguistics. **12**, 157–173 (2024)
32. Wang, Q., Gayed, J.M.: Effectiveness of large language models in automated evaluation of argumentative essays: finetuning vs. zero-shot prompting. Computer Assisted Language Learning. 1–29. https://doi.org/10.1080/09588221.2024.2371395.
33. Latif, E., Zhai, X.: Fine-tuning ChatGPT for automatic scoring. Computers and Education: Artificial Intelligence. **6**, 100210 (2024)
34. Feng, H., Du, S., Zhu, G., Zou, Y., Phua, P.B., Feng, Y., Zhong, H., Shen, Z., Liu, S.: Leveraging Large Language Models for Automated Chinese Essay Scoring. In: Olney, A.M., Chounta, I.-A., Liu, Z., Santos, O.C., and Bittencourt, I.I. (eds.) Artificial Intelligence in Education. pp. 454–467. Springer Cham (2024)
35. OpenAI: o3-mini System Card, https://openai.com/index/o3-mini-system-card/, last accessed 2025/02/19
36. DeepSeek-AI: DeepSeek-R1: Incentivizing Reasoning Capability in LLMs via Reinforcement Learning, https://github.com/deepseek-ai/DeepSeek-R1/blob/main/DeepSeek_R1.pdf, last accessed 2025/02/19
37. Shermis, M.D.: State-of-the-art automated essay scoring: Competition, results, and future directions from a United States demonstration. Assessing Writing. **20**, 53–76 (2014)
38. Yannakoudakis, H., Briscoe, T., Medlock, B.: A New Dataset and Method for Automatically Grading ESOL Texts. In: Lin, D., Matsumoto, Y., and Mihalcea, R. (eds.) Proceedings of the 49th Annual Meeting of the Association for Computational Linguistics: Human Language Technologies. pp. 180–189. Association for Computational Linguistics (2011)